\documentclass[letterpaper]{article} 

\usepackage{aaai2026}
\usepackage{times}  
\usepackage{helvet}  
\usepackage{courier}  
\usepackage[hyphens]{url}  
\usepackage{graphicx} 
\usepackage{amsmath}
\usepackage{amsfonts}

\usepackage{pifont}   
\newcommand{\cmark}{\ding{51}} 
\newcommand{\xmark}{\ding{55}}

\usepackage{natbib}  
\usepackage{caption} 
\usepackage{algorithm}
\usepackage{algorithmic}

\usepackage{newfloat}
\usepackage{listings}
\DeclareCaptionStyle{ruled}{labelfont=normalfont,labelsep=colon,strut=off} 
\floatstyle{ruled}
\newfloat{listing}{tb}{lst}{}
\floatname{listing}{Listing}
\nocopyright 

\title{2DGS-R: Revisiting the Normal Consistency Regularization in \\ 2D Gaussian Splatting}

\author{
    Haofan Ren\textsuperscript{\rm 1}, 
    Qingsong Yan\textsuperscript{\rm 2},
    Ming Lu\textsuperscript{\rm 3},
    Rongfeng Lu\textsuperscript{\rm 1},
    Zunjie Zhu\textsuperscript{\rm 1} \\
}
\affiliations{

    \textsuperscript{\rm 1}Hangzhou Dianzi University
    \textsuperscript{\rm 2}Wuhan University
    \textsuperscript{\rm 3}Intel Labs China\\

}

\usepackage{bibentry}

\begin{document}

\maketitle


\begin{abstract}
Recent advancements in 3D Gaussian Splatting (3DGS) have greatly influenced neural fields, as it enables high-fidelity rendering with impressive visual quality. However, 3DGS has difficulty accurately representing surfaces. In contrast, 2DGS transforms the 3D volume into a collection of 2D planar Gaussian disks. Despite advancements in geometric fidelity, rendering quality remains compromised, highlighting the challenge of achieving both high-quality rendering and precise geometric structures. This indicates that optimizing both geometric and rendering quality in a single training stage is currently unfeasible. To overcome this limitation, we present 2DGS-R, a new method that uses a hierarchical training approach to improve rendering quality while maintaining geometric accuracy. 2DGS-R first trains the original 2D Gaussians with the normal consistency regularization. Then 2DGS-R selects the 2D Gaussians with inadequate rendering quality and applies a novel in-place cloning operation to enhance the 2D Gaussians. Finally, we fine-tune the 2DGS-R model with opacity frozen. Experimental results show that compared to the original 2DGS, our method requires only 1\% more storage and minimal additional training time. Despite this negligible overhead, it achieves high-quality rendering results while preserving fine geometric structures. These findings indicate that our approach effectively balances efficiency with performance, leading to improvements in both visual fidelity and geometric reconstruction accuracy.
\end{abstract}
\section{Introduction}
\label{sec:intro}


\begin{figure}[!t]
    \centering
    \includegraphics[width=\linewidth]{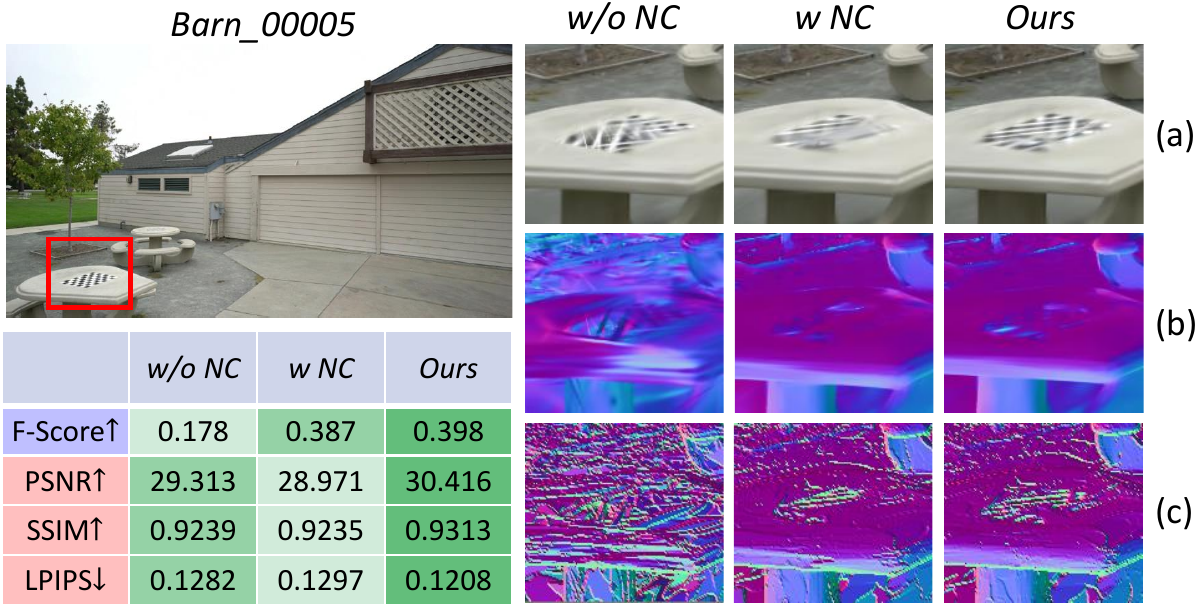}
    \caption{
    (a) Rendered color. (b) Rendered normal. (c) Normal from rendered depth. Our motivation for this work is to achieve high-quality rendering and precise surface reconstruction for 2DGS~\cite{huang20242dgs} at the same time. Normal consistency regularization (NC) means minimizing the difference between (b) and (c).}
    \label{fig:teaser}
\end{figure}

Novel view synthesis (NVS) and 3D surface reconstruction have consistently presented challenges in computer vision and graphics. Recently, 
3D Gaussian Splatting (3DGS)~\cite{kerbl2023gaussian} has emerged as a promising approach for NVS, achieving an effective balance between rendering quality and real-time performance. As a result, it rapidly achieved wide application in diverse fields, including dynamic scene reconstruction~\cite{duan20244d,bae2024ed3dgs,lu2024gagaussian}, autonomous driving~\cite{Zhou_2024_CVPR,zhou2024drivinggaussian}, and SLAM~\cite{yan2024gs,keetha2024splatam,matsuki2024gaussian}. However, the multiview inconsistency of 3D Gaussians lacks a clear definition of surfaces, resulting in difficulties in extracting high-quality meshes.

To obtain better geometry reconstruction, subsequent work like
SuGaR~\cite{guedon2023sugar} forces the 3D Gaussians to align with the surface using geometry regularization. Another type of method, like 2DGS~\cite{huang20242dgs} and GaussianSurfels~\cite{dai2024high}, utilizes the flatten primitive to fit the surface. While these methods enhance the quality of geometric reconstruction, they inevitably result in a slight decrease in rendering quality compared to the original. This observation reflects the ongoing challenge of reconciling rendering quality with geometric accuracy in neural field-based representations.

Among all reconstruction methods based on Gaussian, 2DGS exhibits excellent simplicity and elegance.
Compared with the 3DGS, 2DGS utilizes 2D-to-2D projection with homogeneous coordinates to avoid the unprecise approximation~\cite{zwicker2004perspective}. It improves the quality of geometric reconstruction with a small training time load. We find that geometric regularization, particularly normal consistency, is nearly essential for high-quality geometric reconstruction, but it negatively impacts rendering quality. The results on DTU demonstrate that incorporating normal consistency improves reconstruction accuracy by 46\% compared to cases without it, albeit with a 0.8 dB reduction in PSNR. We propose a novel and effective training strategy that significantly improves rendering quality while preserving geometric reconstruction accuracy. Our key contributions are summarized as follows:
 \begin{itemize}
     \item We conduct extensive experiments to assess the impact of incorporating normal consistency (NC) on the 2DGS attributes. Based on our findings, we propose a hierarchical training strategy that dynamically adjusts different 2DGS attributes throughout the training process.
     \item We analyze the limitations of directly adjusting 2DGS properties and introduce in-place clone densification as an effective complementary strategy to further enhance rendering quality, especially in regions with fine details or abrupt color changes.
     \item By combining the two methods outlined above, we effectively mitigate the trade-off between rendering quality and reconstruction accuracy, with minimal impact on storage and computational load.
 \end{itemize}




\section{Related Work}
\label{sec:related}

\subsection{Novel View Synthesis}

NeRF~\cite{mildenhall2020nerf} leverages a multi-layer perceptron (MLP) for scene representation, encoding both geometry and view-dependent appearance information. Through volume rendering~\cite{max1995optical}, the MLP is optimized with a photometric loss function. Subsequent advancements have focused on optimizing NeRF’s training through feature-grid representations~\cite{yu2021plenoctrees,yu2022plenoxels,muller2022instant,chen2022tensorf} and enhancing rendering speed via baking~\cite{hedman2021baking,reiser2021kilonerf,tang2023delicate}. Besides, NeRF has also been adapted to address anti-aliasing~\cite{barron2023zipnerf,hu2023Tri-MipRF,barron2021mip} and unbounded scenes~\cite{barron2022mip360,zhang2020nerf++}.

More recently, 3D Gaussian splatting~\cite{kerbl2023gaussian} models complex scenes using 3D Gaussians. It achieved outstanding results, with efficient optimization and the capability to render high-resolution images in real-time rendering. Subsequent works improved its storage efficiency~\cite{hac2024,navaneet2023compact3d, Niedermayr_2024_CVPR,wang2024rdogaussian} or rendering efficiency~\cite{jo2024identifying,lee2024gscore}. Moreover, some work~\cite{lu2024scaffold,cheng2024gaussianpro} attempts to improve its rendering quality further. Besides, it also has been extended to surface reconstruction~\cite{Yu2024GOF, guedon2023sugar}.

\subsection{3D reconstruction}

3D reconstruction from multi-view images is a fundamental problem in computer vision. Multi-view stereo methods~\cite{schonberger2016pixelwise,yao2018mvsnet,yu2020fast} often involve intricate, multi-stage processing pipelines. These typically include feature matching, depth estimation, point cloud fusion, and surface reconstruction. Despite their widespread application in academia and industry, these methods are susceptible to artifacts arising from incorrect feature matching and noise introduced at various pipeline stages. In contrast, neural surface reconstruction has leveraged pure deep neural networks to predict surface models directly from multiple image conditions in an end-to-end manner. However, these methods typically involve substantial computational overhead during network inference and require extensively labeled 3D training models, limiting their real-time and practical applicability. While 3DGS~\cite{kerbl2023gaussian} benefits from an explicit scene representation that accelerates training and rendering speed, the absence of well-defined boundaries adversely affects geometric reconstruction quality.

To resolve this challenge, several methods have been introduced, including SuGaR~\cite{guedon2023sugar}, GOF~\cite{Yu2024GOF}, Gaussian Surfels~\cite{dai2024high}, and 2DGS~\cite{huang20242dgs}. 2D Gaussian Splatting (2DGS)~\cite{huang20242dgs} has recently gained attention as a novel technique that simplifies 3D scene representation by converting volumetric data into 2D oriented Gaussian disks. 2DGS offers enhanced geometric reconstruction performance over 3DGS while maintaining efficiency. However, 2DGS exhibits limitations in rendering quality, as reflected by both qualitative and quantitative evaluations. In response to this limitation, we propose a novel training approach for 2DGS that significantly improves rendering quality without compromising geometric accuracy.
\section{Method}
\label{sec:method}

\subsection{Preliminaries}
Given central position $\mu$, a scaling vector $\boldsymbol{S} = (s_{u}, s_{v})$ that governs the covariance of a 2D
Gaussian~\cite{huang20242dgs} and a $3 \times 3$ rotation matrix $\boldsymbol{R} = [\boldsymbol{t}_{u}, \boldsymbol{t}_{v}, \boldsymbol{t}_{w}]$ that defines the orientation of the 2D Gaussian, the transformation between UV space and world space can be expressed as follows:
\begin{equation}
    \boldsymbol{H} = \begin{bmatrix}
        s_{u}\boldsymbol{t}_{u} & s_{v}\boldsymbol{t}_{v} & \boldsymbol{0} & \boldsymbol{p} \\
        0 & 0 & 0 & 1
    \end{bmatrix} = \begin{bmatrix}
        \boldsymbol{RS} & \mu \\
        \boldsymbol{0} & 1 
    \end{bmatrix}
  \label{eq:2dgs_param}
\end{equation}

If $\boldsymbol{W}$ is marked as the transformation matrix from world space to screen space, a homogeneous ray emitted from the camera and passing through pixel (x, y) can be expressed as follows:
\begin{equation}
    \boldsymbol{x} = (xz, yz, z, 1)^\top = \boldsymbol{W}\boldsymbol{H}(u, v,1,1)^\top
  \label{eq:2d_projection}
\end{equation}
where z represents intersection depth. In the rasterization, we input pixel coordinate $(x, y)$ and inquiry intersection in Gaussian's local coordinate. To achieve that, we need to compute the inverse transformation of the projection~\eqref{eq:2d_projection}. The intersection depth z is governed by the constraints of the view-consistent 2D Gaussian. By solving this constraint equation, we can obtain the final result, as described in~\cite{huang20242dgs}: 
\begin{equation}
    u(\boldsymbol{x}) = \frac{\boldsymbol{h}_{u}^{2}\boldsymbol{h}_{v}^{4} - \boldsymbol{h}_{u}^{4}\boldsymbol{h}_{v}^{2}}{\boldsymbol{h}_{u}^{1}\boldsymbol{h}_{v}^{2} - \boldsymbol{h}_{u}^{2}\boldsymbol{h}_{v}^{1}}\quad v(\boldsymbol{x}) = \frac{\boldsymbol{h}_{u}^{4}\boldsymbol{h}_{v}^{1} - \boldsymbol{h}_{u}^{1}\boldsymbol{h}_{v}^{4}}{\boldsymbol{h}_{u}^{1}\boldsymbol{h}_{v}^{2} - \boldsymbol{h}_{u}^{2}\boldsymbol{h}_{v}^{1}} 
  \label{eq:2d_projection_inverse}
\end{equation}
\begin{equation}
    \boldsymbol{h}_{u} = (\boldsymbol{WH})^\top(-1,0,0,x)^\top \quad\boldsymbol{h}_{u} = (\boldsymbol{WH})^\top(0,-1,0,y)^\top 
  \label{eq:2d_projection_inverse_2}
\end{equation}
where $(x,y)$ is the pixel coordinate and $\boldsymbol{h}_{u}^{i}$, $\boldsymbol{h}_{v}^{i}$ represent the i-th parameter of the vector.

The loss function in 2DGS is as follows:
\begin{equation}
    \mathcal{L} = \mathcal{L}_c + \alpha\mathcal{L}_d + \beta\mathcal{L}_n
    \label{eq:2dgs_loss}
\end{equation}

The depth distortion $\mathcal{L}_d$ and normal consistency $\mathcal{L}_n$ regularization are as follows:

\begin{equation}
    \mathcal{L}_d=\sum_{i, j} \omega_i \omega_j\left|z_i-z_j\right|
    \label{eq:depth_distortion}
\end{equation}
where $\omega_i$ means the weight of the $i$-th intersection and $z_{i}$ is the depth of the intersection points.

\begin{equation}
    \mathcal{L}_n=\sum_i \omega_i\left(1-\mathbf{n}_i^{\mathrm{T}} \boldsymbol{N}\right)
    \label{eq:normal_consistency}
\end{equation}
where $i$ indexes the intersected splats along the ray, 
$\omega$ signifies the blending weight at the intersection point, $\boldsymbol{n}_{i}$ indicates the normal of the splat facing the camera, and $\boldsymbol{N}$ is the normal estimate from the gradient of the depth map. $\mathbf{p}_s$ means the intersection between surfels and rays omitted from the pixel.

\begin{equation}
    \mathbf{N}(x, y)=\frac{\nabla_x \mathbf{p}_s \times \nabla_y \mathbf{p}_s}{\left|\nabla_x \mathbf{p}_s \times \nabla_y \mathbf{p}_s\right|}
\end{equation}

Once the training is finished, the depth map can be derived using the following formula to obtain the reconstructed result. Afterward, 2DGS utilizes TSDF fusion for mesh extraction with depth.

\begin{equation}
    z_{\text {mean}}=\sum_i \omega_i z_i /(\sum_i \omega_i+\epsilon)
    \label{eq:depth_map}
\end{equation}

where $\omega_i=T_i \alpha_i \hat{\mathcal{G}}_i(\mathbf{u}(\mathbf{x}))$ represent the weight contribution of $i$-th Gaussian and $T_i=\prod_{j=1}^{i-1}(1-\alpha_j \hat{\mathcal{G}}_j(\mathbf{u}(\mathbf{x})))$ means its visibility.

Besides, 2DGS also provides a more robust depth value extraction method. It determines the median depth as the greatest "visible" depth, using $T_{i} = 0.5$ as the threshold between surface and free space. The last Gaussian is selected if a ray's accumulated alpha remains below 0.5.

\begin{equation}
    z_{\text {median}}=\max\{z_{i}|T_{i} > 0.5\}
    \label{eq:median_depth_map}
\end{equation}

\begin{table}[]
    \begin{tabular}{cc|ccc|c}
    \hline
    $\alpha$ & $\beta$ & PSNR$\uparrow$  & SSIM$\uparrow$ & LPIPS$\downarrow$ & F-score$\uparrow$ \\ \hline
    0      & 0     & 24.54 & 0.838 & 0.200 & 0.15    \\
    100/10 & 0     & \textbf{23.56} & 0.817 & 0.225 & 0.14    \\
    100/10 & 0.05  & 23.08 & 0.809 & 0.235 & 0.32    \\
    0      & 0.05  & 24.30 & 0.836 & 0.203 & 0.33    \\
    0      & 0.1   & 24.07 & 0.832 & 0.207 & \textbf{0.36}    \\
    0      & 0.3   & 22.74 & 0.804 & 0.243 & 0.32    \\ \hline
    \end{tabular}
    \caption{\textbf{The effect of different regularization terms}. The results from the TnT dataset. 100/10 in $\alpha$ denotes the settings for 360-degree and large-scale scenes as described in 2DGS, respectively.}
    \label{tab:alpha_beta}
\end{table}

\subsection{The Impact of Normal Consistency}
\label{sec:analysis}

As shown in Table~\ref{tab:alpha_beta}, depth distortion $\mathcal{L}_d$ exhibits negligible impact on both rendering quality and geometric reconstruction. Normal consistency(NC) $\mathcal{L}_n$ plays a more important role in 2DGS surface reconstruction. It facilitates the local alignment of all 2D splats with the underlying surface. In fact, it is a very strong geometric constraint that affects all properties of 2DGS except for spherical harmonics. Increasing the weight of $\mathcal{L}_d$
improves geometric reconstruction quality but degrades rendering quality. Moreover, if $\mathcal{L}_d$ becomes too dominant, both geometric reconstruction and rendering quality deteriorate.

First, we analyze the impact of NC loss on the spatial coverage of 2DGS. Each Gaussian has the following attributes $\{\mu,\Sigma,\alpha, \text{SHs}\}$: central position $\mu$, covariance matrix $\Sigma$, opacity $\alpha$ and SH coefficient. We primarily focus on $\boldsymbol{S}$ and $\alpha$, as it is challenging to analyze the other attributes using a fair and quantitative method. In particular, we utilize $K_{a}$ to characterize $\boldsymbol{S}=(s_{u}, s_{v})$, which represents the spatial coverage.

\begin{equation}
    K_{a} = s_{u} \times s_{v}
\end{equation}

\begin{figure}[!t]
    \centering
    \includegraphics[width=\linewidth]{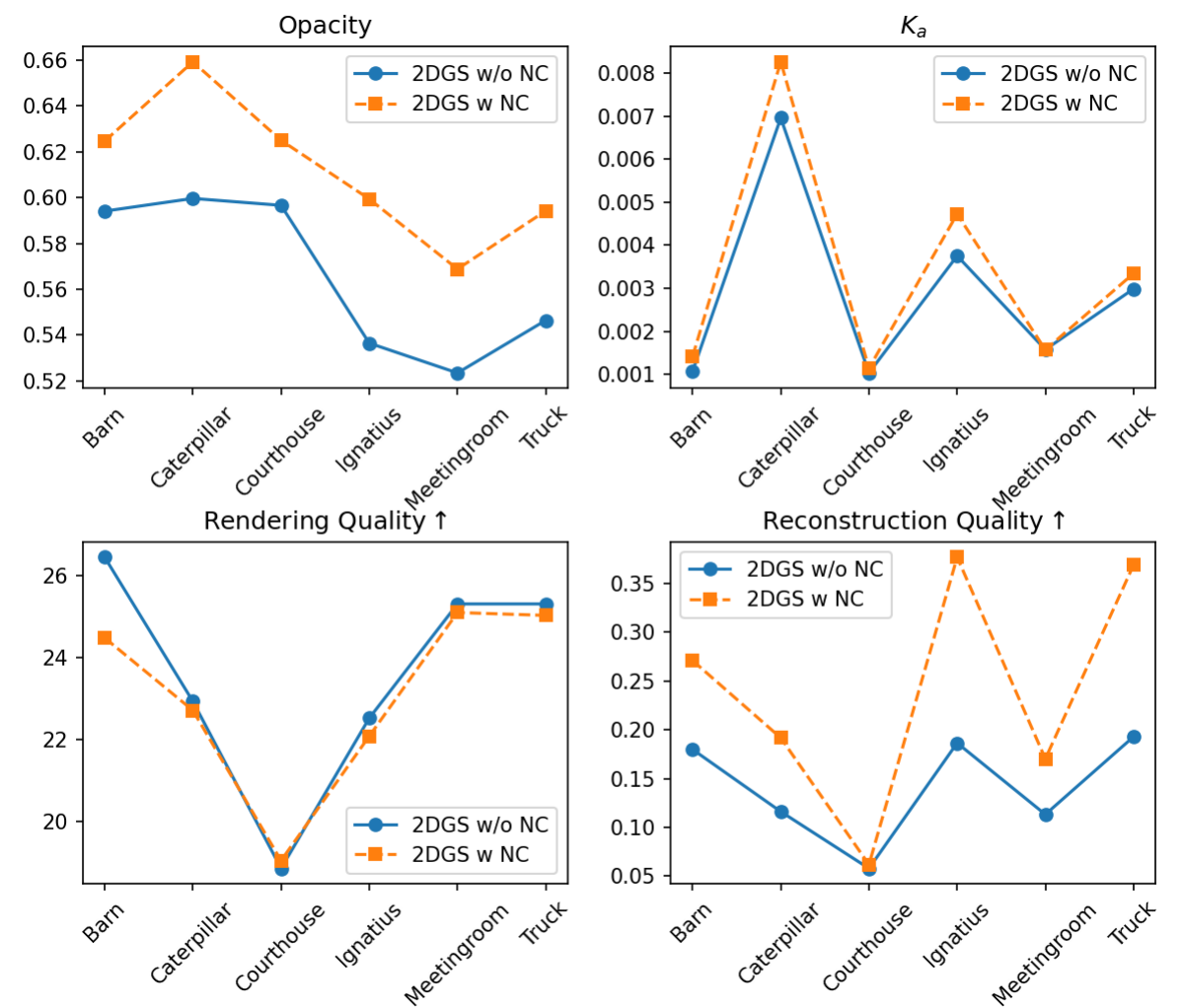}
    \caption{
    Each scene's average value of 2DGS on the Tanks and Temples Dataset~\cite{Knapitsch2017}. The introduction of NC increases $K_{a}$ and opacity. The larger the $K_a$ and opacity, the better the reconstruction quality, while the rendering quality will deteriorate.}
    \label{fig:2dgs_attributes}
\end{figure}

As shown in Figure~\ref{fig:2dgs_attributes}, the introduction of NC increases both $K_{a}$ and opacity $\alpha$. Moreover, the larger 
$K_{a}$ and opacity $\alpha$, the higher the reconstruction quality, but the lower the rendering quality. Figure~\ref{fig:2dgs_attributes} also shows that the properties required for optimal rendering quality differ from those needed for the best reconstruction results in 2DGS. This suggests that resolving the conflict between rendering quality and geometric reconstruction may be impossible. However, the traditional texture mapping~\cite{zhou2014color,lee2020texturefusion,ha2021normalfusion}, demonstrates that enhancing rendering quality is possible without altering the underlying geometry. Naturally, this observation leads to a simple yet effective approach to addressing the conflict. We prioritize training the spatial distribution of the 2DGS first, followed by fine-tuning the spherical harmonic coefficients to optimize rendering quality. However, the rendering results are still inferior to those of 2DGS without NC. 


\twocolumn[{
    \renewcommand\twocolumn[1][]{#1}
    \maketitle
    \centering
    \begin{minipage}{1.00\textwidth}
        \centering 
        \includegraphics[clip=False, width=\linewidth]{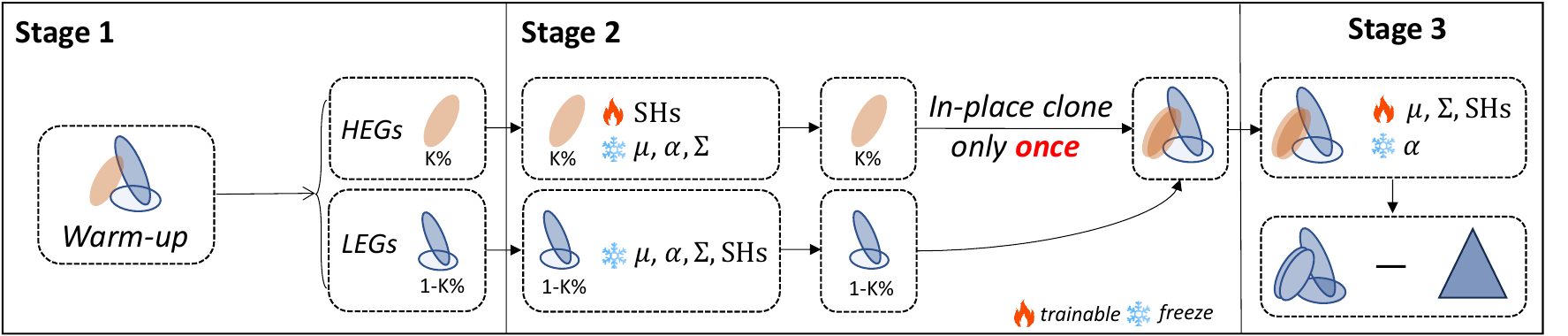}
    \end{minipage}
    \captionsetup{type=figure}
    \captionof{figure}{Our training process is divided into three stages. To illustrate this, we use a pure blue triangle as the ground truth example. Each ellipse denotes a 2D Gaussian, and the color variations reflect the evolution of appearance as training progresses.
    }
    \label{fig:pipeline}
}]

\begin{figure}[htbp]
    \centering
    \includegraphics[width=\linewidth]{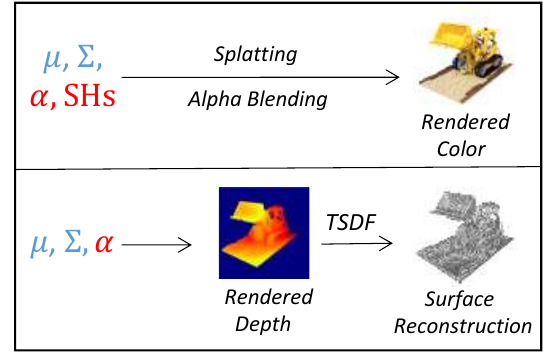}
    \caption{$\mu$ and $\Sigma$ represent the geometric properties of the Gaussians, while $\alpha$ and SHs are used to represent the appearance model of the Gaussians.}
    \label{fig:gs_attributes}
\end{figure}

As shown in Figure~\ref{fig:gs_attributes}, we analyze how different Gaussian attributes affect rendering quality and geometric reconstruction. We observe that as training progresses, 
$\{\mu,\Sigma\}$ no longer fluctuate as drastically as they did in the early stages of training. This implies that, with 
$\{\mu,\Sigma\}$ fixed, we can achieve a balance between rendering quality and geometric reconstruction by adjusting $\{\alpha, \text{SHs}\}$. In this case, $\alpha$ plays a crucial role in determining both geometry and rendering.


\begin{figure}[!t]
    \centering
    \includegraphics[width=\linewidth]{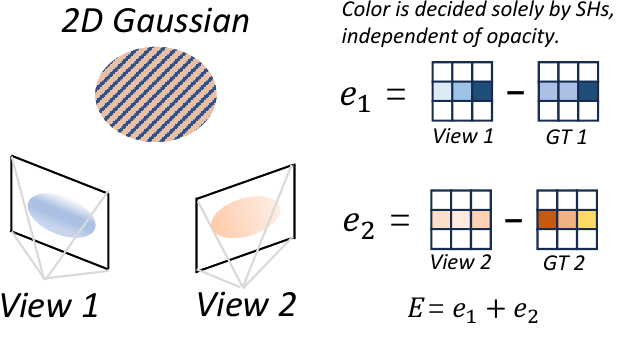}
    \caption{
    We take a single Gaussian and two perspectives as examples to demonstrate how to calculate $E_{i}$.
    }
    \label{fig:gserror}
\end{figure}

\subsection{Stage 1 and Stage 2}
\label{sec:finetune_sh}

During Stage 1 and Stage 2, we focus on fine-tuning the appearance while preserving the geometry. In Stage 1, we train 2DGS with normal consistency, resulting in a model with well-distributed Gaussians in space but suboptimal appearance. Upon completion of this stage, we obtain $N$ Gaussians. We then compute the color error between each Gaussian and the corresponding ground-truth pixel across all training views, accumulating the errors to form a per-Gaussian error metric. The detailed computation is provided in Eq.~\ref{eq:gs_error}, and a simple illustrative example for computing $E_i$ is shown in Figure~\ref{fig:gserror}.

\begin{equation}
    E_{i} = \sum_{v=1}^{\mathcal{V}}\sum_{i=1}^{N} 
    \mathbb{L}(i,v) \cdot \mathbb{K}(i,v) \cdot \left| C_{i}^{v}- C_{gt}^{v} \right|
    \label{eq:gs_error}
\end{equation}

\begin{equation}
\mathbb{L}(i,v)=
\begin{cases}
    1, & G_i \in \mathcal{F}(v) \\
    0, & G_i \notin \mathcal{F}(v)
\end{cases}
\end{equation}

\begin{equation}
\mathbb{K}(i,v)=
\begin{cases}
    1, & C_{gt}^{v} \in \mathcal{S}(i, v) \\
    0, & C_{gt}^{v} \notin \mathcal{S}(i, v)
\end{cases}
\end{equation}

where $\mathcal{V}$ means the training-set views, $C_{i}^{v}$ means the color of $i$-th Gaussian under training-set view $v$, $C_{gt}^{v}$ represents the ground truth color corresponding to the pixel positions occupied by $i$-th Gaussian $G_{i}$ under view $v$.  $\mathcal{F}(v)$ means the frustum of view $v$. $\mathcal{S}(i, v)$ means the set of pixel position occupied by $i$-th Gaussian $G_{i}$ under view $v$.

\begin{figure}[!t]
    \centering
    \includegraphics[width=\linewidth]{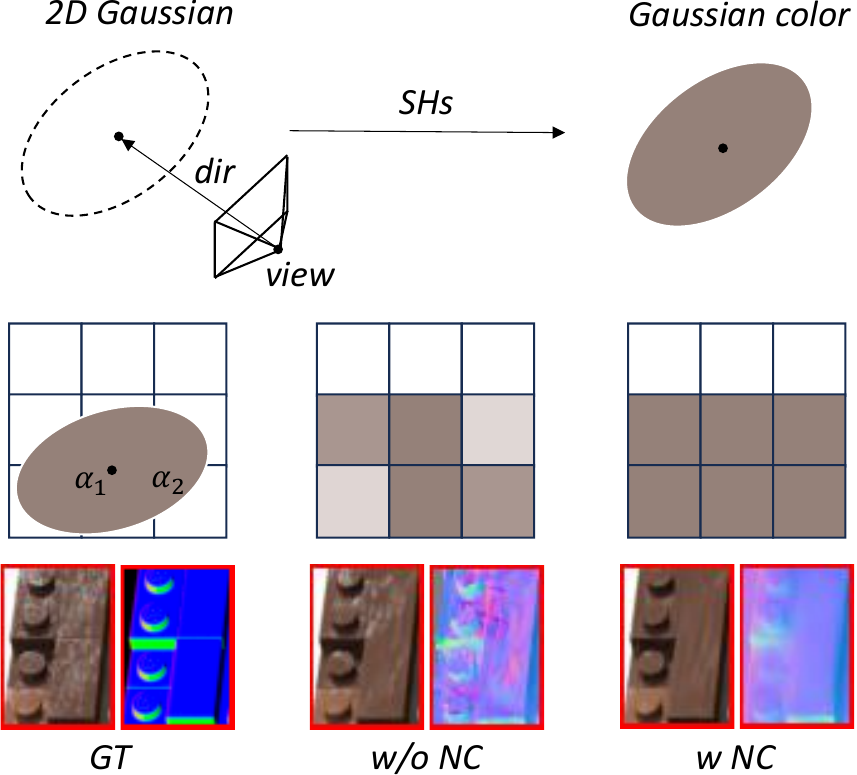}
    \vspace{-1.5 em}
    \caption{
    The top image illustrates the rendering process of a single Gaussian, while the bottom showcases the impact on the rendered color and normal after introducing NC.
    }
    \label{fig:nce}
\end{figure}

As illustrated in Figure~\ref{fig:nce}, for each observed viewpoint, the color of a Gaussian is computed via spherical harmonics (SH) using the direction vector from the camera to the Gaussian center. The Gaussian is then splatted onto the image plane, where the final pixel color is influenced by both the Gaussian’s color and its opacity. We design $E_i$ to highlight regions with significant color variation but relatively flat geometry—such as the mottled area in the lower-left corner of Figure~\ref{fig:nce}. In such regions, we expect $E_i$ to yield higher values. To this end, $E_i$ is computed without considering the effect of $\alpha$ (opacity), and it is not normalized by the number of occupied pixels. This design enhances its correlation with $K_a$. For $i$-th Gaussian, we obtain corresponding $E_{i}$. For the $i$-th Gaussian, we compute the corresponding error score $E_i$. All Gaussians are then sorted in descending order based on their $E_i$ values. We select the top $K\%$ as high-error Gaussians (HEGs), and designate the rest as low-error Gaussians (LEGs). During Stage 2 (Figure~\ref{fig:pipeline}), we freeze all attributes of the LEGs, while for HEGs, only the SH coefficients remain trainable. Subsequently, both groups are jointly trained without applying normal consistency.

\subsection{Stage 3}
\label{sec:in-place_clone}

In Stage 3, the optimization objective shifts to jointly fine-tuning appearance and geometry to achieve higher rendering fidelity and structural accuracy.
We consider a typical scenario where the pixel coverage of fine structures is smaller than the $K_{a}$ of a single 2D Gaussian. In such cases, merely adjusting ${\alpha, \text{SHs}}$ is insufficient to improve rendering quality, as illustrated in the bottom of Figure~\ref{fig:nce}. When color varies rapidly over small spatial regions, large Gaussians with high opacity fail to capture these variations accurately, resulting in rendering artifacts.

To address this limitation, we propose a novel targeted enhancement strategy that adaptively increases modeling capacity in challenging regions. Specifically, we introduce a small number of additional Gaussians in fine-grained or mottled areas that are not well represented by the existing set. However, two key challenges arise: (1) where to insert new Gaussians, and (2) which attributes should be fine-tuned to enhance rendering while preserving geometry.

To solve the first challenge, we establish a guiding principle: improve rendering quality while minimizing geometric distortion. Based on this principle, we choose to add as few Gaussians as possible. Our key insight is that high-error Gaussians (HEGs) inherently identify regions that are difficult to reconstruct accurately. Due to their large coverage and small quantity, HEGs are ideal insertion points. Therefore, we perform in-place cloning of HEGs to generate new Gaussians specifically at these critical locations.

To address the second challenge, we carefully select the attributes to optimize. Since SHs influence appearance without affecting geometry, we fine-tune SH coefficients for the newly added Gaussians. In addition, we optimize $\Sigma$ to allow these Gaussians to better conform to complex spatial color patterns, such as mottled regions. However, opacity ($\alpha$) affects both appearance and depth, and adjusting it can distort geometry. Hence, we adopt a freeze-opacity (FO) strategy during this stage. This design reflects another core contribution of our method: a decoupled fine-tuning mechanism that separates appearance adaptation from geometric refinement.

In essence, our three-stage pipeline is designed to progressively disentangle and resolve appearance-geometry conflicts. From geometry-aware initialization to error-driven refinement and targeted enhancement, each stage plays a distinct role. This strategy not only improves rendering quality in visually complex regions, but also maintains global geometric consistency with minimal additional overhead.

\begin{table*}[htbp]
  \centering
  \small
  \resizebox{\textwidth}{!}{%
  \begin{tabular}{c|cccc|cccc|cccc}
    \hline
    Dataset &\multicolumn{4}{c|}{TnT} & \multicolumn{4}{c|}{NeRF Synthetic} & \multicolumn{4}{c}{DTU}\\
    \hline
    Method & F-Score$\uparrow$ & PSNR$\uparrow$ & SSIM$\uparrow$ & LPIPS$\downarrow$ &  
    F-Score$\uparrow$ & PSNR$\uparrow$ & SSIM$\uparrow$ & LPIPS$\downarrow$ & 
    CD$\downarrow$ & PSNR$\uparrow$ & SSIM$\uparrow$ & LPIPS$\downarrow$  \\
    \hline
    NeuS & 0.38 & 24.58 & -- & -- & -- & -- & -- & -- & 0.84 & 31.97 & 0.840 & -- \\
    3DGS & -- & -- & -- & -- & -- & -- & -- & -- & 4.03 & 32.91 & 0.943 & 0.092 \\
    Sugar & -- & -- & -- & -- & -- & -- & -- & -- & 1.24 & 32.76 & 0.942 & 0.094 \\
    GSDF & -- & -- & -- & -- & -- & -- & -- & -- & 0.80 & 33.65 & 0.948 & 0.092 \\
    GOF\textsuperscript{*} & 0.46 & 23.06 & 0.845 & 0.177 & 0.908 & 33.28 & 0.969 & 0.031 & 0.75 & 35.11 & 0.949 & 0.130 \\
    PGSR\textsuperscript{*} & 0.40  & 23.79 & 0.851 & 0.160 & 0.907 & 31.84 & 0.964 & 0.036 & 0.56 & 33.78 & 0.945 & 0.147 \\
    SVRaster & 0.40 & 23.04 & -- & 0.144 & -- & -- & -- & -- & -- & -- & -- & -- \\
    \hline
    2DGS\textsuperscript{*} & \textbf{0.33} & 24.30 & 0.836 & 0.203 & 0.905 & 31.89 & 0.965 & 0.038 & 0.77 & 34.78 & 0.939 & 0.165 \\
    2DGS(w/o NC) & 0.15 & 24.54 & 0.838 & 0.200 & 0.874 & \textbf{33.06} & \textbf{0.968} & \textbf{0.033} & 1.40 & 35.54 & 0.941 & 0.163 \\
    2DGS-R & \textbf{0.33} & \textbf{24.73} & \textbf{0.841} & \textbf{0.197} & \textbf{0.911} & 32.92 & 0.967 & 0.034 & \textbf{0.74} & \textbf{35.69} & \textbf{0.943} & \textbf{0.159} \\
    \hline
    
  \end{tabular}
   }%
  \caption{\textbf{Quantitative comparison on three datasets.} The rendering quality on DTU is derived from the training set, while the rendering quality on TnT and NeRF Synthetic datasets is evaluated from novel view synthesis. \textsuperscript{*} indicates results reproduced using the official source code. The F-Score reported in PGSR original paper is 0.52, which primarily due to using a smaller voxel size and leads to OOM under the 256GB memory. We set the voxel size to match that of the 2DGS.
  }
  \label{tab:benchmark}
\end{table*}


\section{Experiments and Results}
\label{sec:experiments}


\subsection{Datasets and Metrics}
 We assess the performance of our method using both synthetic and real datasets, specifically the NeRF-Synthetic~\cite{mildenhall2020nerf}, Tanks \& Temples (TNT)~\cite{Knapitsch2017} and DTU~\cite{jensen2014large} dataset. We evaluate geometric reconstruction and rendering quality on these three datasets at the same time.
For training and evaluation, we keep the resolution of the NeRF Synthetic dataset and we downsample the resolution of the DTU dataset to 1/2. For TnT dataset, we downsample the original image to half its resolution.

For rendering quality, we use PSNR, SSIM~\cite{wang2004ssim} and LPIPS~\cite{zhang2018perceptual} for quantitative comparison on NeRF Synthetic, TnT and DTU datasets. Following the strategy of 2DGS~\cite{huang20242dgs}, we evaluated PSNR on the training set for the DTU dataset. To evaluate the quality of geometry reconstruction, we adopt different metrics depending on the characteristics of each dataset. On the DTU dataset, we leverage the availability of high-quality ground-truth meshes and calculate the bidirectional Chamfer Distance (CD) between them and the reconstructed mesh extracted from the predicted TSDF. For the NeRF Synthetic and TnT datasets, where ground-truth surface geometry is not directly accessible, we instead compute the F1-score, which provides a meaningful measure of geometric accuracy under the given constraints.

\subsection{Implementation and Baseline}
We compare with NeuS~\cite{wang2021neus}, 3DGS~\cite{bae2024ed3dgs}, Sugar~\cite{guedon2023sugar}, GSDF~\cite{yu2024gsdf}, GOF~\cite{Yu2024GOF}, PGSR~\cite{chen2024pgsr}, SVRaster~\cite{sun2025sparse} and 2DGS~\cite{huang20242dgs}. Training stage 1 keeps the same strategy as the original 2DGS. Specifically, we increase the SH degree every 1000 steps until the maximum degree is reached. The number of iterations in the training stage is set to 30k, 10k, and 20k, respectively. We set $K=1$ for all datasets, resulting in only a 1\% increase in storage compared to 2DGS. All experiments are conducted on an RTX 4090 GPU.

\begin{figure}[htbp]
\centering
\includegraphics[width=\linewidth]{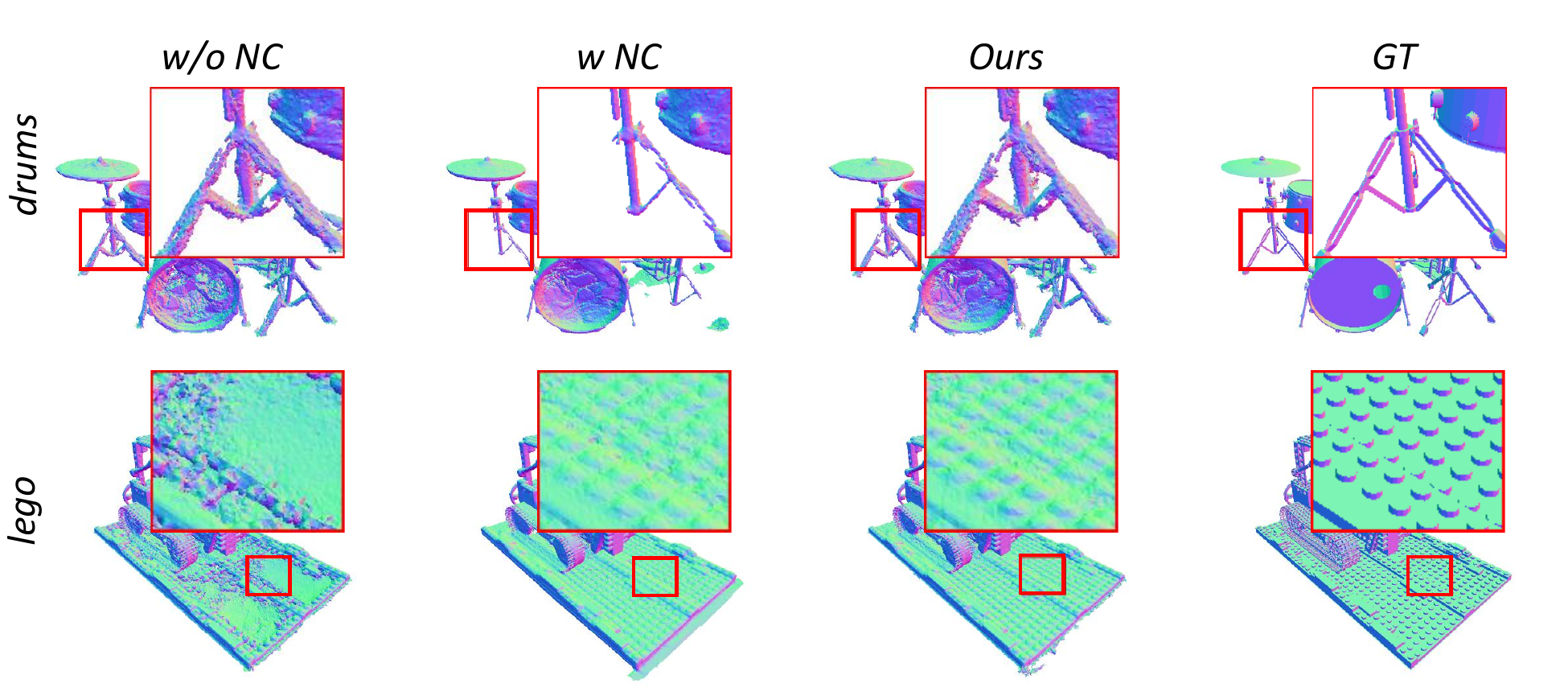}
\vspace{-1.5 em}
\caption{\textbf{Visual comparison of mesh normal results on Synthetic NeRF dataset.}
In the \textit{drums} case, our method can reconstruct a more complete scene. Our method can recover better details in \textit{lego} case.
}

\label{fig:ns_recon}
\end{figure}

\begin{figure}[htbp]
    \centering
    \includegraphics[width=\linewidth]{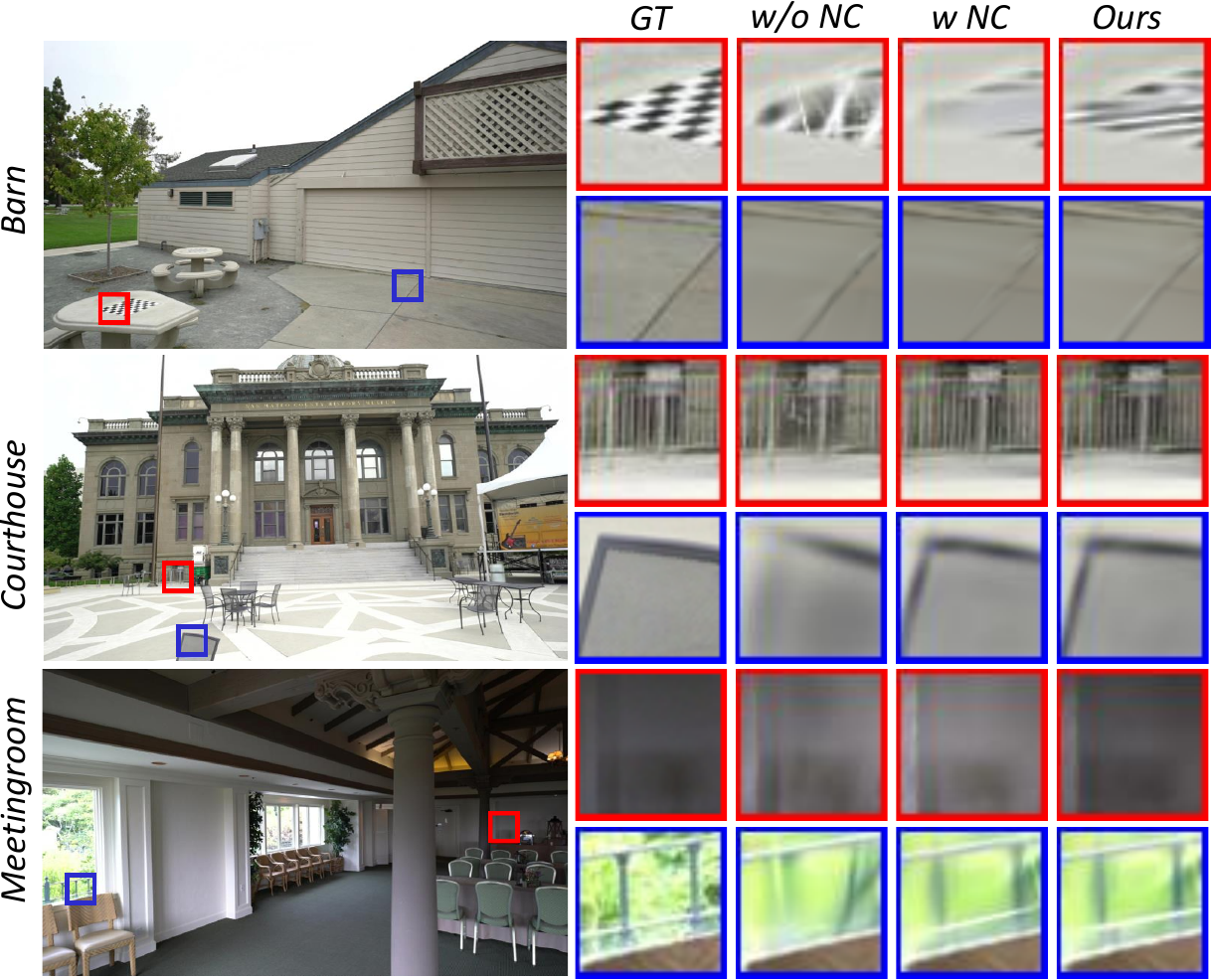}
    \caption{\textbf{Visual comparison of novel view synthesis on TnT dataset.} Our method can better reconstruct the details of the scene in the \textit{Barn}.
    As shown in \textit{Meetingroom}, our approach helps reduce artifacts in low-texture areas.
    }
    \label{fig:tnt_nvs}
\end{figure}

\begin{figure}[!htbp]
    \centering
    \includegraphics[width=\linewidth]{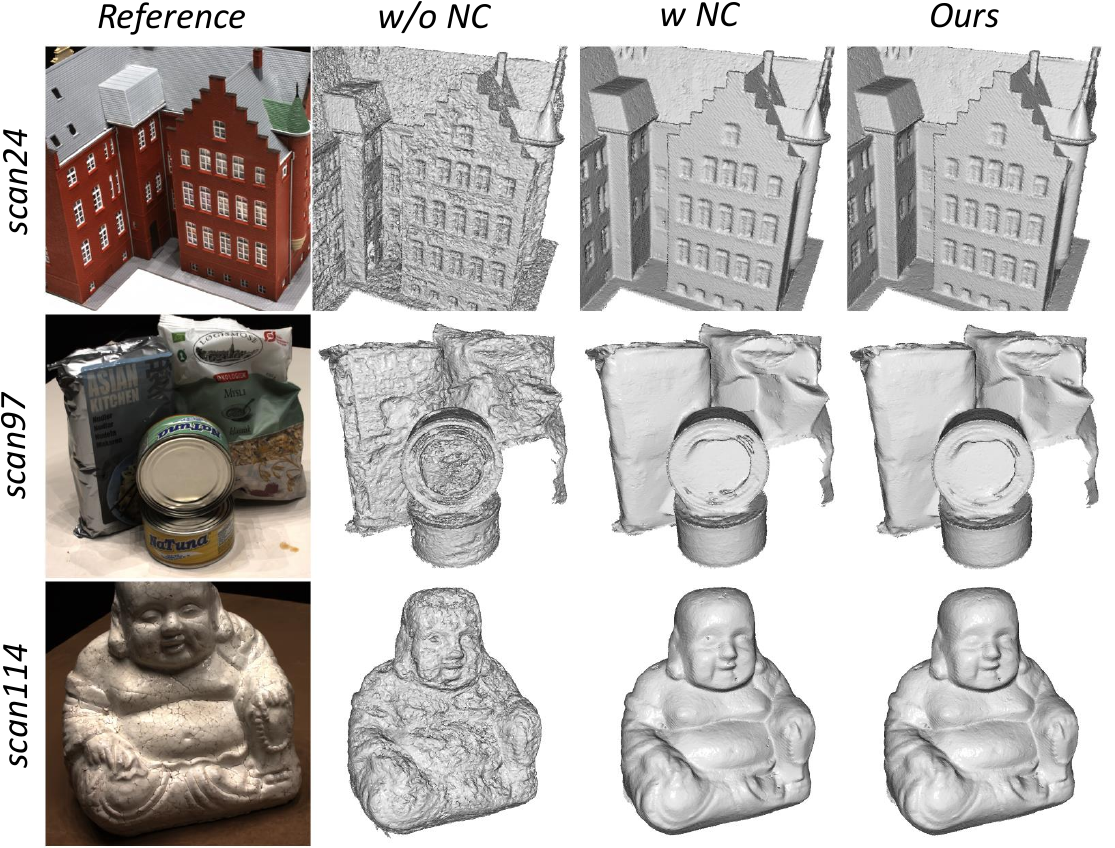}
    \caption{\textbf{Visual comparison of geometric reconstruction on DTU dataset.}
    }
    \label{fig:dtu_recon}
\end{figure}

\subsection{Results and Comparision}

\paragraph{Rendering Quality \& Geometry Reconstruction}

\begin{figure}[!htbp]
    \centering
    \includegraphics[width=\linewidth]{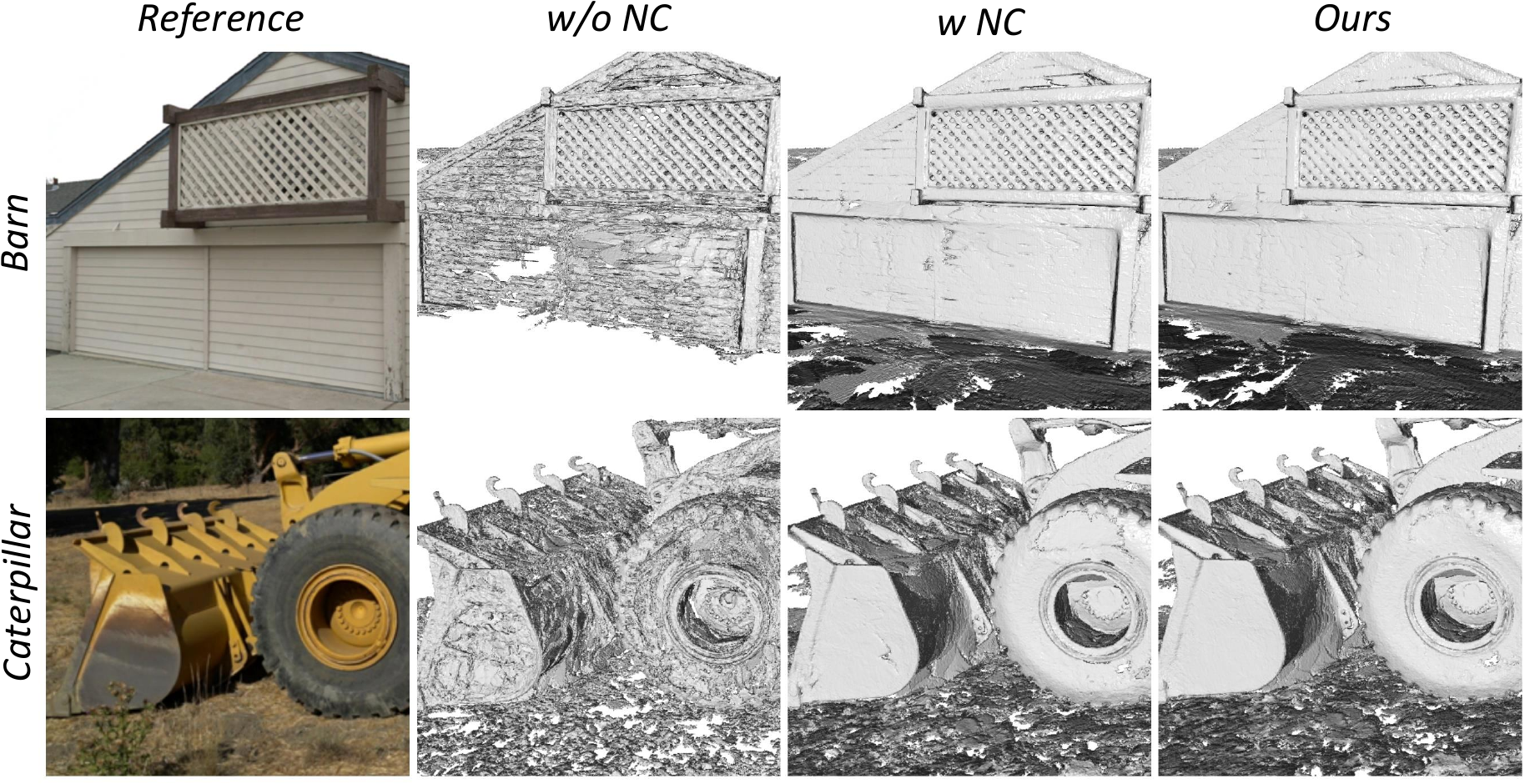}
    \caption{\textbf{Visual comparison of geometric reconstruction on TnT dataset.} Compared to directly introducing NC, our method produces smoother reconstruction results.
    }
    \label{fig:tnt_mesh}
\end{figure}

We compare rendering quality (with training-set views) on the DTU dataset and novel view synthesis (NVS) on the NeRF Synthetic and TnT datasets. NVS results in TnT are shown in Figure~\ref{fig:tnt_nvs}. Geometric reconstruction is evaluated on all three datasets. The results as shown in Table~\ref{tab:benchmark} and Figure~\ref{fig:ns_recon},~\ref{fig:dtu_recon},~\ref{fig:tnt_mesh}. Our method consistently achieves competitive performance across both rendering and reconstruction tasks.

As shown in Figure~\ref{fig:tnt_nvs}, our method successfully recovers high-frequency details and generates more accurate color representations, particularly in low-texture areas. Additionally, Figure~\ref{fig:ns_recon} demonstrates that our approach reconstructs cleaner and more complete meshes, featuring sharper and better-defined edges.

As illustrated in Figure~\ref{fig:dtu_recon}, our method yields more accurate reconstructions, especially along challenging regions such as the edges of the can and the area surrounding the Buddha’s left eye. Similarly, in Figure~\ref{fig:tnt_mesh}, our approach effectively avoids noticeable breakages along the wall boundaries and delivers a smoother reconstruction of the excavator’s bucket.

\section{Discussion and Conclusion}
\label{sec:discussion}

We first conducted comprehensive ablation studies, including the influence of $K$, the impact of each training stage, the effect of fine-tuning settings, and the use of alternative operations to replace in-place cloning. We report the quantitative evaluation results in Table~\ref{tab:ablation_K},~\ref{tab:ablation_freeze},~\ref{tab:ab_module},~\ref{tab:absgs}.

\subsection{Ablation}

\paragraph{The impact of K}
We validate the impact of different $K$ values on rendering quality and reconstruction quality on the TnT dataset. As shown in Table~\ref{tab:ablation_K}, despite the increase in $K$, there is no noticeable improvement in rendering quality and geometric reconstruction.

\begin{table}[htbp]
  \centering
  \footnotesize
  \begin{tabular}{c|cccc}
    \hline
    &F-Score$\uparrow$ & PSNR$\uparrow$ & SSIM$\uparrow$ & LPIPS$\downarrow$ \\
    \hline
    K=1 & 0.33 & 24.73 & 0.841 & 0.197 \\
    K=3 & 0.33 & 24.70 & 0.841 & 0.197 \\
    K=5 & 0.33 & 24.71 & 0.841 & 0.196 \\
    K=10 & 0.33 & 24.72 & 0.841 & 0.195 \\
    K=30 & 0.33 & 24.75 & 0.842 & 0.193 \\
    K=50 & \textbf{0.34} & \textbf{24.78} & \textbf{0.843} & \textbf{0.191} \\
    \hline
  \end{tabular}
  \caption{\textbf{Quantitative ablation study on K.}}
  \label{tab:ablation_K}
\end{table}

\paragraph{The impact of finetune attributes}
We experiment with different combinations of Gaussian attribute fine-tuning in training Stage 3 to observe their impact. Comparison is shown in Table~\ref{tab:ablation_freeze}. It can be seen that the impact on the F-score is not significant, but freezing different attributes causes a noticeable difference in rendering quality.

\paragraph{The impact of each module}

We evaluate the effectiveness of each module from the TnT dataset. As shown in Table~\ref{tab:ab_module}(I), applying NC in 2DGS as a basic baseline. (II) While employing in-place clone operation, the improvement in PSNR is quite significant, whereas the F-score drops drastically. (III) By incorporating the freezing opacity operation, both PSNR and F-score show a slight improvement. (IV) Resuming NC in Stage 3 leads to a slight decrease in PSNR, but a significant improvement in F-score.

\paragraph{Replace the in-place clone with AbsGS}
We replace the in-place clone densification strategy with an alternative strategy, AbsGS~\cite{ye2024absgs}, and the results are shown in Table~\ref{tab:absgs}. Since the original AbsGS is based on 3DGS, directly using its hyperparameters on 2DGS named AbsGS-D is not appropriate. Therefore, we also fine-tuned the hyperparameters for 2DGS named AbsGS-A. It can be observed that after parameter tuning, the rendering quality improves, but the geometric quality deteriorates. This is typically due to the fact that gradient-based densification strategies are highly sensitive to the loss function and often require multiple iterations, which can cause fluctuations in the number of Gaussians. As a result, it is usually necessary to adjust the hyperparameter settings for different scenes. 

\begin{table}[htbp]
  \centering
  \footnotesize
  \begin{tabular}{c|cccc}
    \hline
    Freeze/Finetune & F-Score$\uparrow$ & PSNR$\uparrow$ & SSIM$\uparrow$ & LPIPS$\downarrow$ \\
    \hline
    (,)/($\mu$,$\alpha$,$\Sigma$,SHs) & 0.33 & 24.61 & 0.840 & 0.198 \\
    ($\alpha$,SHs)/($\mu$,$\alpha$,$\Sigma$) & 0.33 & 24.52 & 0.838 & 0.201\\
    (SHs)/($\mu$,$\alpha$,$\Sigma$) & 0.33 & 24.50 & 0.837 & 0.201 \\
    ($\alpha$)/($\mu$,$\alpha$,$\Sigma$,SHs) & 0.33 & \textbf{24.73} & \textbf{0.841} & \textbf{0.197} \\
    \hline
  \end{tabular}
  \caption{\textbf{Ablation study on freeze/finetune of TNT dataset.} The strategy of freezing opacity yields the best results compared with other strategies.}
  \label{tab:ablation_freeze}
\end{table}

\begin{table}[htbp]
  \centering
  \footnotesize
  \begin{tabular}{c|ccc|c|c}
    \hline
     Methods & Clone & FO & RNC & F-Score$\uparrow$ & PSNR$\uparrow$ \\
    \hline
    (I) & \xmark & \xmark & \xmark & 0.33 & 24.30 \\
    (II) & \cmark & \xmark & \xmark & 0.24 & 24.84 \\
    (III) & \cmark & \cmark & \xmark & 0.26 & 24.89 \\
    (IV) & \cmark & \cmark & \cmark & 0.33 & 24.73 \\
    \hline
  \end{tabular}
  \caption{\textbf{Ablation study on the 6 scenes of TNT dataset}. "Clone" means the in-place clone operation. "FO" means the freeze opacity of 2D Gaussians in Stage 3. "RNC" means resuming the NC in Stage 3.}
  \label{tab:ab_module}
\end{table}

\begin{table}[htbp]
  \centering
  \small
  \resizebox{\linewidth}{!}{\begin{tabular}{c|cc|cc|cc}
    \hline
    &\multicolumn{2}{c|}{TNT} & \multicolumn{2}{c|}{NeRF Synthetic} &  \multicolumn{2}{c}{DTU} \\
    Method & F-Score$\uparrow$ & PSNR$\uparrow$ & F-Score$\uparrow$ & PSNR$\uparrow$ & CD$\downarrow$ & PSNR$\uparrow$ \\
    \hline
    2DGS(w/o NC) & 0.15 & 24.54 & 0.874 & 33.06 & 1.40 & 35.54 \\
    2DGS(w NC) & 0.33 & 24.30 & 0.905 & 31.89 & 0.77 & 34.78 \\
    \hline
    2DGS-R(AbsGS-D) & 0.19 & 23.19 & 0.899 & 32.53 & 0.95 & 33.53 \\
    2DGS-R(AbsGS-A) & 0.08 & 24.26 & 0.904 & \textbf{33.15} & 0.90 & 34.57 \\
    2DGS-R(ours) & \textbf{0.33} & \textbf{24.73} & \textbf{0.911} & 32.92 & \textbf{0.74} & \textbf{35.69} \\
    \hline
  \end{tabular}}
  \caption{\textbf{Quantitative comparison of rendering quality and reconstruction quality.} AbsGS-D means using the original hyperparameters, while AbsGS-A means that the original hyperparameters
have been adjusted with more sensitive thresholds for size and positional gradient.}
  \label{tab:absgs}
\end{table}

\subsection{Conclusion}
In this work, we study the effect of introducing Normal Consistency (NC) into Gaussian attributes. While NC improves reconstruction accuracy, it can degrade rendering quality. To balance this trade-off, we propose a simple yet effective multi-stage training strategy that progressively refines different attributes, mitigating the conflict between reconstruction and rendering with only a slight increase in training time.

\bibliography{aaai2026}

\end{document}